\pdfoutput=1

\documentclass[11pt]{article}

\usepackage[]{EMNLP2023}
\usepackage{times}
\usepackage{latexsym}
\usepackage{booktabs}
\usepackage{graphicx}
\usepackage{caption}
\usepackage{subcaption}
\usepackage{multirow}
\usepackage{amsmath}

\usepackage[normalem]{ulem}

\usepackage[T1]{fontenc}

\usepackage[utf8]{inputenc}

\usepackage{microtype}
\usepackage{multirow}
\usepackage{color,soul}
\usepackage{times}
\usepackage[T1]{fontenc}
\usepackage{latexsym}
\usepackage{booktabs}

\usepackage[T1]{fontenc}

\usepackage[utf8]{inputenc}
\usepackage{soul}
\usepackage{color}

\usepackage{microtype}

\usepackage{inconsolata}

\definecolor{hlcoherence}{cmyk}{0.0, 0.31, 0.22, 0.09}
\DeclareRobustCommand{\hlcoherence}[1]{{\sethlcolor{hlcoherence}\hl{#1}}}
\definecolor{hlconsistency}{cmyk}{0.57, 0.07, 0.0, 0.09}
\DeclareRobustCommand{\hlconsistency}[1]{{\sethlcolor{hlconsistency}\hl{#1}}}
\definecolor{hlfluency}{cmyk}{0.14, 0.25, 0.0, 0.09}
\DeclareRobustCommand{\hlfluency}[1]{{\sethlcolor{hlfluency}\hl{#1}}}
\definecolor{hlrelevance}{cmyk}{0, 0, 0.0, 0.29}
\DeclareRobustCommand{\hlrelevance}[1]{{\sethlcolor{hlrelevance}\hl{#1}}}

\title{The Impact of Preference Agreement in Reinforcement Learning from Human Feedback: A Case Study in Summarization}

\author{Sian Gooding \\
  Google Research \\
  \texttt{sgooding@google.com} \\\And
  Hassan Mansoor \\
   Google Research \\
  \texttt{hassan@google.com} \\}

\begin{document}
\maketitle
\begin{abstract}
Reinforcement Learning from Human Feedback (RLHF) can be used to capture complex and nuanced properties of text generation quality. As a result, the task of text summarization has been identified as a good candidate for this process. In this paper, we explore how preference agreement impacts the efficacy of RLHF for summarization. We show that sampling human preferences to include a range of annotator agreement results in (1) higher accuracy reward models and (2) alters the characteristics of quality captured. We additionally show improvements in downstream generation when using a reward model trained with a range of preference agreements. Our contributions have implications for the design of synthetic datasets as well as the importance of considering quality differentials in comparison-based data.


\end{abstract}

\section{Introduction}

\begin{table}[t]
\small 
\centering
\begin{tabular}{llc} \toprule
    & \textit{Summary}                                                                                                                                                                                                          & \multicolumn{1}{l}{\textit{Votes}} \\ \toprule
$1$ (a) & \begin{tabular}[c]{@{}l@{}}In my road trip across the US, \\ should I try to revisit childhood \\ memories, or should I go places \\ that I've never been?\end{tabular}                                                   & {11}                                       \\ \midrule
$1$ (b) & \begin{tabular}[c]{@{}l@{}}What's the best route for \\ a road trip across the US? \\ Should I go places that mean \\ something to me from my \\ childhood, or should I try to \\ go places I've never seen?\end{tabular} & {11}                                       \\ \midrule \midrule
$2$ (a) & \begin{tabular}[c]{@{}l@{}}Dating a 37 year old married \\ man who is still living with his\\  wife.  \hl{This is not what friends} \\  \hl{see when they look at me.} \\ Help me please.\end{tabular}                                & {0}                                       \\ \midrule
$2$ (b) & \begin{tabular}[c]{@{}l@{}}I am dating a married man\\  who is separated from his \\ wife but still lives with her, \\ none of my friends support \\ me in this situation.\end{tabular}                                   & {22}         \\ \bottomrule                              
\end{tabular}
\caption{Two summary comparisons from the \citet{NEURIPS2020_1f89885d} dataset -- the votes column shows the absolute number of preferences for each summary. $1$ (a) is compared with $1$ (b) and elicits low agreement as each summary is voted for $11$ times; $2$ (a) is compared with $2$ (b) and yields complete agreement.}
\label{tab:comparisons}
\end{table}
 \begin{figure*}
    \centering
    \begin{subfigure}{0.5\textwidth}
        \centering
        \includegraphics[width=0.8\linewidth]{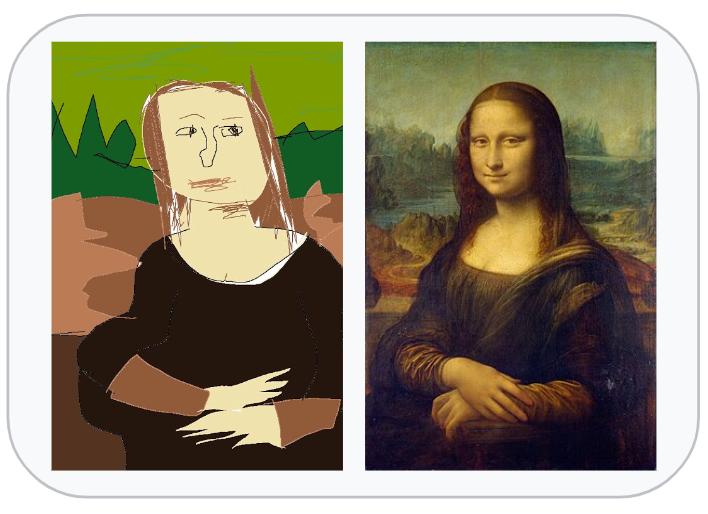}
        \caption{}
        \label{fig:subfiga}
    \end{subfigure}%
    \begin{subfigure}{0.5\textwidth}
        \centering
        \includegraphics[width=0.8\linewidth]{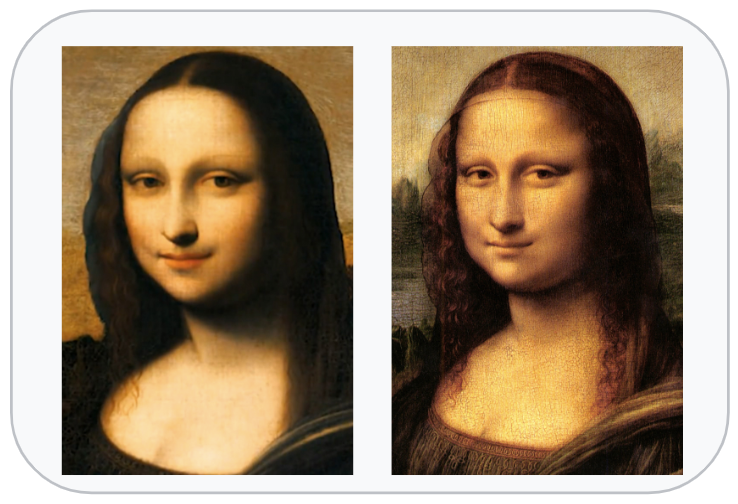}
        \caption{}
       
    \end{subfigure}
    \caption{Two pairs of images showing imitations of the Mona Lisa painting -- when collecting annotations for which rendition is better quality, it is likely that (a) would elicit higher agreement due to the clear quality differential}
     \label{fig:mona_lisa}
\end{figure*}

Language models have exhibited impressive capabilities for text generation. However, what makes a given output appropriate is task-dependent and often subjective. As a result, automated metrics often fail to capture the nuanced aspects of quality for targeted text generation. In order to capture the subtleties of high quality text generation, RLHF focuses on learning a reward function, from human preferences, which can capture complex, nuanced characteristics of text which are difficult to superficially encode via coarse metrics. This technique has been employed in the task of text summarization, whereby such preferences are used to train a reward model which can then be harnessed in a reinforcement learning framework. 

However, not all comparisons are equal, both in perceived difficulty and in informativeness for training a reward model. For instance, Table \ref{tab:comparisons} shows two comparisons with their preference annotations from the \citet{NEURIPS2020_1f89885d} dataset. The first comparison, between summaries $1$ (a) and $1$ (b), elicits low agreement, as an equal number of annotators ($11$) have chosen each as the best generation. However, in the second example there is clear agreement with a unanimous vote for  $2$ (b). The question is, what is the difference between these two sets of summaries that leads to one decision having much higher agreement than the other? Further to this, which comparison is most effective when training a reward model.

In this paper, we extend the work of \citet{NEURIPS2020_1f89885d} by investigating how annotator agreement can be used to gain insights into the difficulty of summary comparisons. We address the following research questions: \\

\begin{enumerate}
    \item How does the diversity of preference agreements, within the training data, impact the accuracy of reward models?
    \item Do the characteristics of summary quality captured by a reward model vary according to the agreement of input preferences?
    \item What are the downstream consequences of training reward models with differing agreement rates?
    \end{enumerate}
More broadly, we consider whether the paradigm of comparison based evaluation benefits from the aim of maximising of agreement. 


\section{Background}
It is widely acknowledged that summarization is difficult to collect human annotation for due to the range of possible solutions introducing subjectivity. However, the framing of summary evaluation as a comparison task introduces an additional axis with which to consider agreement, which is a feature of the comparison itself. Namely, the \textit{quality differential} of the two examples presented. 

For instance, Figure \ref{fig:mona_lisa} presents two pairs of images depicting imitations of the Mona Lisa painting. Hypothetically, when collecting annotations for which rendition is better quality, it is likely that one set of examples would evoke higher agreement due to the clear quality differential. In fact, research has shown that having clear category differences reduces cognitive load and results in increased labelling efficiency \citep{sarkar2016setwise}. For instance, agreement when ranking complex words can be improved by ensuring there are a spread of difficulties represented \cite{gooding-etal-2019-comparative}.  

In our work, we argue that trends in preference agreement are highly influenced by the quality differential. As such, aiming to maximise agreement may not be beneficial as it results in comparisons containing superficial learning opportunities. 

\paragraph{Modelling agreement} The recognition and interpretation of subjective language has been a long-standing focus in the field of Natural Language Processing, with tasks like sentiment analysis and emotion detection being inherently subjective \cite{10.1162/0891201041850885,ovesdotter-alm-2011-subjective}. \citet{ovesdotter-alm-2011-subjective} argues that striving for a single definitive `ground truth' in subjective tasks is impractical and unnecessary, suggesting that understanding annotator subjectivity is more valuable than minimizing annotation variability.

Annotator disagreements can be categorized into random variation and systematic disagreement \cite{krippendorff2011computing}. Several studies have explored ways to utilize annotation disagreement in model training. \citet{prabhakaran-etal-2012-statistical} explored applying higher cost for errors made on unanimous annotations to decrease the penalty of mis-labeling inputs with higher disagreement. \citet{fornaciari-etal-2021-beyond} introduced an auxiliary task for annotator label distributions in a multi-task model, even enhancing performance in less subjective tasks like part-of-speech tagging. While \citet{davani-etal-2022-dealing} use a multi-task architecture using a shared representation to model annotator disagreements, rather than using it in loss function, their approach models several annotators’ labels as multiple tasks and obtains their disagreement. 

However, there is little research on agreement with respect to the quality differential of presented generations. 

\paragraph{Agreement in summarization} Human feedback does not always provide a gold standard for summarization when the task is not clearly defined. It has been established that linguistically trained, expert raters, provide the gold standard in summarization evaluation and the reliability of non-experts has been repeatedly questioned \cite{lloret2018challenging}. For instance, it has been found that crowd workers should not be used to evaluate summary quality because of a non-correlation with experts \cite{gillick2010non, fabbri2021summeval}. Furthermore, even for expert annotations mediation meetings are necessary to assure reliability \citep{iskender2021reliability}. In short, evaluating the quality of a summary is not straightforward.

Across papers, guidelines provided to annotators on what constitutes a good summary have a high degree of variation. For instance, \citet{howcroft2020twenty} found over $200$ variations in terminology when analysing annotator guidelines. Plus, the annotation interface used can influence the agreement of annotators when judging summary comparisons \cite{gooding-etal-2023-study}. Regardless of the type of summary evaluation conducted, the task is difficult due to the subjectivity involved
\cite{fiori2014innovative}: first, because of the lack of agreement on the quality criteria; secondly, because of the subjectivity of assessing
the given criteria; and thirdly, because of the amount of effort required to evaluate summaries can be very time consuming.

When considering summary agreement from the perspective of preferences, \citet{NEURIPS2020_1f89885d} collect a substantial dataset of human comparisons using both researchers and crowd-sourced annotators. The agreement rate among researchers varied across different comparison scenarios. For the most challenging comparisons, which involved comparing two high-temperature samples from a single RL policy, the agreement rate was approximately 65\% -- it should be noted that chance agreement for this comparison task would be 50\%. On the other hand, the easiest comparisons, which involved comparing a high-temperature sample from a supervised baseline to the human reference summary, had an agreement rate of around 80\%. When researchers discussed the comparisons among themselves, the agreement rate increased to $\sim$95\%. The work emphasises that the presence of substantial noise in the annotations is due to the difficulty and subjectivity of the comparisons. Overall, the agreement rate between the annotators in the entire corpus was 72\%. To improve this rate, the paper suggests using the modal output from $3$ annotators, which can increase the agreement to 77\%. 

\begin{figure*}[t]
    \centering
    \includegraphics[width=\textwidth]{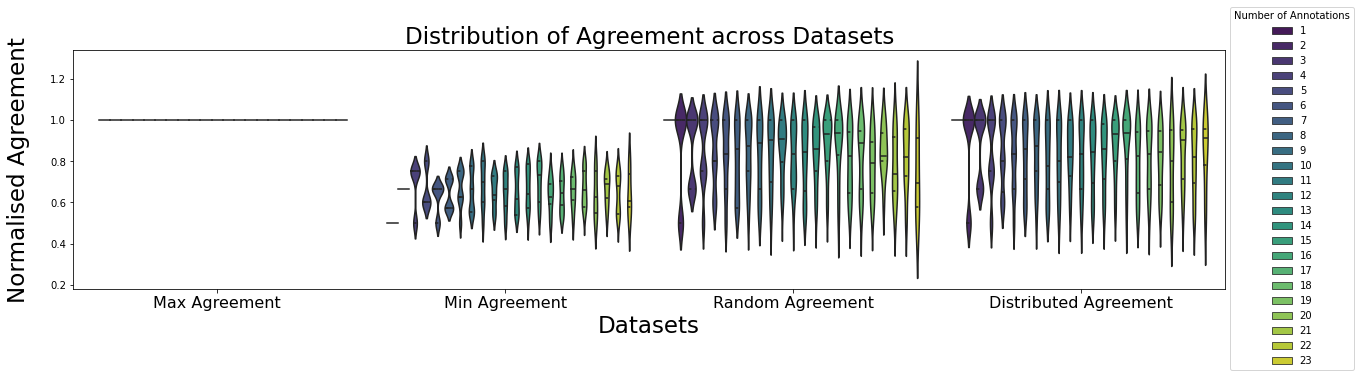}
    \caption{Violin plots showing the distribution of agreement within each dataset as well as the number of repetitions for samples}
    \label{fig:distribution}
\end{figure*}
\paragraph{RLHF for summarization} Deep Reinforcement Learning from Human Preferences \cite{NIPS2017_d5e2c0ad} has increasingly been used in the context of text generation, as well as to align the behaviour of large language models \cite{NEURIPS2020_1f89885d,NEURIPS2022_b1efde53}. This is achieved by the training of a reward model (RM), also known as a preference model, which can be calibrated using human preferences. For text generation, the objective is to develop a model capable of taking a sequence of text as input and returning a scalar reward as output. This reward should represent the likelihood that humans would prefer this generation for the given task, thereby encoding a measure of generation quality.

For the task of summarization, \citet{NEURIPS2020_1f89885d} show that human preference data can be used to improve the quality of model output. This is achieved by training a RM to optimise for human preferences, instead of relying on traditional evaluation measures like R\textsc{ouge} \cite{lin2004rouge}. This reward model can then be harnessed in a RLHF loop when fine-tuning.  

\citet{https://doi.org/10.48550/arxiv.2204.05862} use preference modeling and RLHF for fine-tuning language models for the purpose of serving as helpful and harmless assistants. The authors specifically mention summarization as an example of a helpful task. They found that there was poor average agreement between Anthropic\footnote{\url{https://www.anthropic.com/}} researchers and the data obtained from crowd sourcing. Furthermore, they discovered that agreement between authors and raters was not a reliable indicator for assessing the overall quality of the conversation.

\section{Experimental setup}
In our study, we train four reward models using training input containing differing agreement diversity. To create the training data, we produce datasets containing comparisons sampled according to their agreement as well prioritising samples with a high number of annotations. We observe the differences of trained models using three techniques. Firstly, we observe the accuracy of reward models on a held out comparison dataset. We measure how the score of trained models correlates with features of summary quality and measure the downstream quality of summaries produced when integrating the reward model into an RLHF framework. 

\subsection{Agreement datasets} \label{sec-datasets}
We use an existing dataset collected by \citet{NEURIPS2020_1f89885d} which contains $64,832$ human judgements. Each judgement represents a preference between two generated summaries for an article from the TL;DR summarization dataset \cite{volske2017tl}. We filter the dataset for instances that have been annotated by more than one annotator in order to measure pairwise agreement. The summaries provided for comparison within the dataset are Transformer decoders in the style of GPT-3. 

We create four datasets each containing $2000$ instances which are sampled according to their agreement levels. Figure \ref{fig:distribution} shows the distribution of agreement within each dataset and includes a key showing the number of repetitions present for each sample. 

We create four datasets containing different agreement levels. The first dataset, denoted as \textsc{max}, comprises of the top $2000$ instances within the dataset that have the highest agreement and number of repetitions. The second dataset, labeled \textsc{min}, consists of $2000$ examples with the lowest agreement rates and highest number of repetitions. For the third dataset, named \textsc{dist}, we select comparisons that represent a balanced distribution of agreement rates observed in the original dataset. Lastly, we randomly sample $2000$ comparisons to form the fourth dataset, referred to as \textsc{rand}. The test set contains a total of $1267$ instances, which have been randomly sampled and are not included at train time. 

\subsubsection{Reward model}
The reward models used in our experiments are T5-XXL with 13B parameters, both encoder and decoder components have $24$ transformer blocks and $64$ attention heads.  The encoder input is the concatenation of the context and generation tokens interleaved with the context and generation prefixes, and the decoder input is the scoring tokens.

The RM is trained on a dataset of comparisons between two model outputs on the same input and use a cross-entropy loss, with the comparisons as labels. The difference in rewards represents the log odds that one response will be preferred to the other by a human rater. Loss$(\theta)$ is therefore:
\begin{equation*}
\small
-\frac{1}{{K \choose 2}}E_{(x,y_w,y_l)\sim D}[\log (\sigma (r_\theta(x,y_w) - r_\theta(x,y_l)))]
\end{equation*}
 where the loss function for the reward model is the scalar output of the reward model for context $x$ and generations $y$ with parameters $\theta$ and $y(w)$ is the preferred completion out of the pair of $y_w$ and $y_l$ and $D$ is the dataset of human comparison. 

\subsubsection{RLHF framework}
We adopt the experimental setup from \citet{NEURIPS2020_1f89885d} for our study. However, due to the nature of the policy gradient methods used, which involve computing an estimator of the policy gradient and using a stochastic gradient ascent algorithm, there is a tendency for excessively large policy updates that can be detrimental. Therefore, we impose constraints on the policy update to mitigate this issue.
\section{Results}
\begin{figure}[t]
  \centering
  \includegraphics[width=\columnwidth]{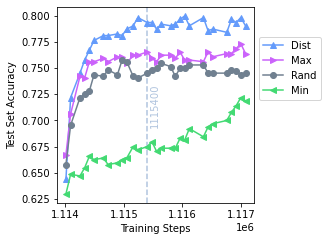}
  \caption{Test set accuracy plotted against the number of training steps for the datasets}
  \label{fig:accuracy_graph}
\end{figure}


Figure \ref{fig:accuracy_graph} illustrates the accuracy improvement of reward models with increasing training steps. Initially, all training regimens show significant accuracy improvement, except for the dataset with low agreement (\textsc{min}), which improves at a slower rate. Notably, the distributed data setting surpasses the other datasets in terms of test set accuracy, demonstrating faster improvement and achieving the highest performance. There is a clear distinction in performance between the dataset with low agreement and all other settings, highlighting the importance of aiming to increase agreement during the data collection process. The \textsc{dist} dataset contains a range of agreement levels and the resulting reward model attains the highest accuracy on the test set. This emphasises the impact of a curriculum-like learning approach whereby the model has the opportunity to learn from both easy and challenging examples.

\subsection{SummEval Quality Correlations}
\begin{figure}[t]
    \centering
    \includegraphics[width=\columnwidth]{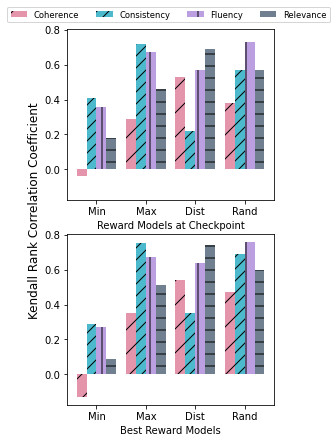}
    \caption{Graphs showing the Kendall Rank correlations between the reward model output and SummEval quality measures}
    \label{fig:kendal_rank_graph}
\end{figure}
\begin{table}[t]
\small
 \centering
\begin{tabular}{llllll}
\toprule
\rotatebox[origin=c]{90}{\textcolor[HTML]{000000}{\textit{}}} &
&
  \rotatebox[origin=c]{60}{\textcolor[HTML]{000000}{\small\textit{{\hlcoherence{Coherence}}}}} &
  \rotatebox[origin=c]{60}{\textcolor[HTML]{000000}{\small{\phantom{-}\textit{\hlconsistency{Consistency}}}}} &
  \rotatebox[origin=c]{60}{\textcolor[HTML]{000000}{\small{\textit{\hlfluency{Fluency}}}}} &
  \rotatebox[origin=c]{60}{\textcolor[HTML]{000000}{\small\textit{{\hlrelevance{Relevance}}}}} \\ \midrule
\rotatebox[origin=c]{60}{\textcolor[HTML]{000000}{}} &
 \textsc{min} &
  \textcolor[HTML]{000000}{{\ul{-0.04}}} &
  \textcolor[HTML]{000000}{\phantom{-}0.41} &
  \textcolor[HTML]{000000}{\phantom{-}\ul{0.36}} &
  \textcolor[HTML]{000000}{\phantom{-}\ul{0.18}} \\
\rotatebox[origin=c]{90}{\textcolor[HTML]{000000}{}} &
  \textsc{max}&
  \textcolor[HTML]{000000}{\phantom{-}0.29} &
  \textcolor[HTML]{000000}{\phantom{-}\textbf{0.72}} &
  \textcolor[HTML]{000000}{\phantom{-}0.67} &
  \textcolor[HTML]{000000}{\phantom{-}0.46} \\
\rotatebox[origin=c]{90}{\textcolor[HTML]{000000}{}} &
  \textsc{dist}&
  \textcolor[HTML]{000000}{\phantom{-}\textbf{0.53}} &
  \textcolor[HTML]{000000}{\phantom{-}\ul{0.22}} &
  \textcolor[HTML]{000000}{\phantom{-}0.57} &
  \textcolor[HTML]{000000}{\phantom{-}\textbf{0.69}} \\ 
\multirow{-4}{*}{\rotatebox[origin=c]{90}{\textcolor[HTML]{000000}{\textit{Checkpoint}}}} &
 \textsc{rand} &
  \textcolor[HTML]{000000}{\phantom{-}0.38} &
  \textcolor[HTML]{000000}{\phantom{-}0.57} &
  \textcolor[HTML]{000000}{\phantom{-}\textbf{0.73}} &
  \textcolor[HTML]{000000}{\phantom{-}0.57} \\  \midrule
\rotatebox[origin=c]{90}{\textcolor[HTML]{000000}{}} &
 {Min} &
  \textcolor[HTML]{000000}{-\ul{0.13}} &
  \textcolor[HTML]{000000}{\phantom{-}\ul{0.29}} &
  \textcolor[HTML]{000000}{\phantom{-}\ul{0.27}} &
  \textcolor[HTML]{000000}{\phantom{-}\ul{0.09}} \\
\rotatebox[origin=c]{90}{\textcolor[HTML]{000000}{}} &
 \textsc{max}&
  \textcolor[HTML]{000000}{\phantom{-}0.35} &
  \textcolor[HTML]{000000}{\phantom{-}\textbf{0.75}} &
  \textcolor[HTML]{000000}{\phantom{-}0.67} &
  \textcolor[HTML]{000000}{\phantom{-}0.51} \\
\rotatebox[origin=c]{90}{\textcolor[HTML]{000000}{}} &
  \textsc{dist}&
  \textcolor[HTML]{000000}{\phantom{-}\textbf{0.54}} &
  \textcolor[HTML]{000000}{\phantom{-}0.35} &
  \textcolor[HTML]{000000}{\phantom{-}0.64} &
  \textcolor[HTML]{000000}{\phantom{-}\textbf{0.74}} \\
\multirow{-4}{*}{\rotatebox[origin=c]{90}{\textcolor[HTML]{000000}{\textit{Best}}}} &
  \textsc{Rand} &
  \textcolor[HTML]{000000}{\phantom{-}0.47} &
  \textcolor[HTML]{000000}{\phantom{-}0.69} &
  \textcolor[HTML]{000000}{\phantom{-}\textbf{0.76}} &
  \textcolor[HTML]{000000}{\phantom{-}0.60} \\  \midrule
\rotatebox[origin=c]{90}{\textcolor[HTML]{000000}{}} &
 \textsc{rouge-}$1$ &
  \textcolor[HTML]{000000}{\phantom{-}0.25} &
  \textcolor[HTML]{000000}{\phantom{-}0.53} &
  \textcolor[HTML]{000000}{\phantom{-}0.52} &
  \textcolor[HTML]{000000}{\phantom{-}0.41} \\ 
  & \textsc{rouge-}$2$ &
  \textcolor[HTML]{000000}{\phantom{-}0.16} &
  \textcolor[HTML]{000000}{\phantom{-}0.59} &
  \textcolor[HTML]{000000}{\phantom{-}0.48} &
  \textcolor[HTML]{000000}{\phantom{-}0.29} \\ 
 &  \textsc{rouge-l} &
  \textcolor[HTML]{000000}{\phantom{-}0.07} &
  \textcolor[HTML]{000000}{\phantom{-}0.15} &
  \textcolor[HTML]{000000}{\phantom{-}0.29} &
  \textcolor[HTML]{000000}{\phantom{-}0.24} \\ \bottomrule
\end{tabular}
\caption{Kendall rank correlations between reward model output and SummEval quality measures}
\label{tab:kendal_rank_tab}
\end{table}

\begin{figure*}[t]
  \centering
  \begin{subfigure}[b]{0.3\textwidth}
    \includegraphics[width=\textwidth]{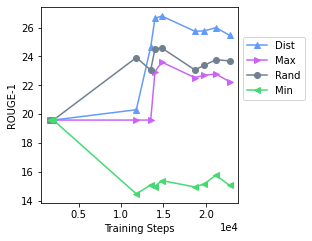}
    \caption{\textsc{rouge-}$1$}
    \label{fig:figure1}
  \end{subfigure}
  \hfill
  \begin{subfigure}[b]{0.3\textwidth}
    \includegraphics[width=\textwidth]{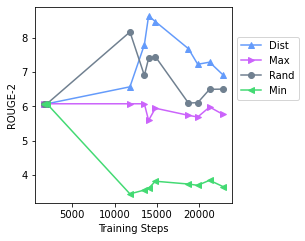}
    \caption{\textsc{rouge-}$2$}
    \label{fig:figure2}
  \end{subfigure}
  \hfill
  \begin{subfigure}[b]{0.3\textwidth}
    \includegraphics[width=\textwidth]{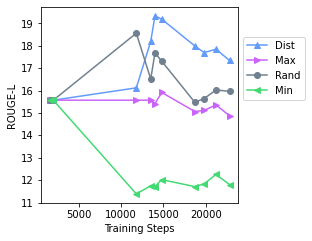}
    \caption{\textsc{rouge-L}}
    \label{fig:figure3}
  \end{subfigure}

  \caption{Graphs showing the improvement in scores over training steps for different training data using the following metrics (a) \textsc{rouge-}$1$, (b) \textsc{rouge-}$2$, and (c) \textsc{rouge-}L}
  \label{fig:overall}
\end{figure*}

To evaluate whether characteristics of summary quality are captured by the reward models we use the SummEval dataset \cite{fabbri2021summeval}. The dataset contains human judgments of model-generated summaries based on  news articles contained in the CNN/Dailymail dataset \cite{hermann2015teaching}. Each summary is annotated by both expert judges and crowd-source workers. We use the annotations provided by experts due to crowd-sourced ratings being deemed less reliable. Each generated summary is evaluated according to the following four criteria: \\
\begin{itemize}
    \item \textit{\hlcoherence{Coherence}}: well-structured and well-organized;
    \item \textit{\hlconsistency{Consistency}}: the factual alignment between the summary and the summarized source;
    \item  \textit{\hlfluency{Fluency}}: the quality of individual sentences;
    \item \textit{\hlrelevance{Relevance}}: selection of important content from the source.
\end{itemize}

\begin{table*}[t]
\fbox{%
\begin{tabular}{lll}
\multirow{3}{*}{\includegraphics[width=2cm]{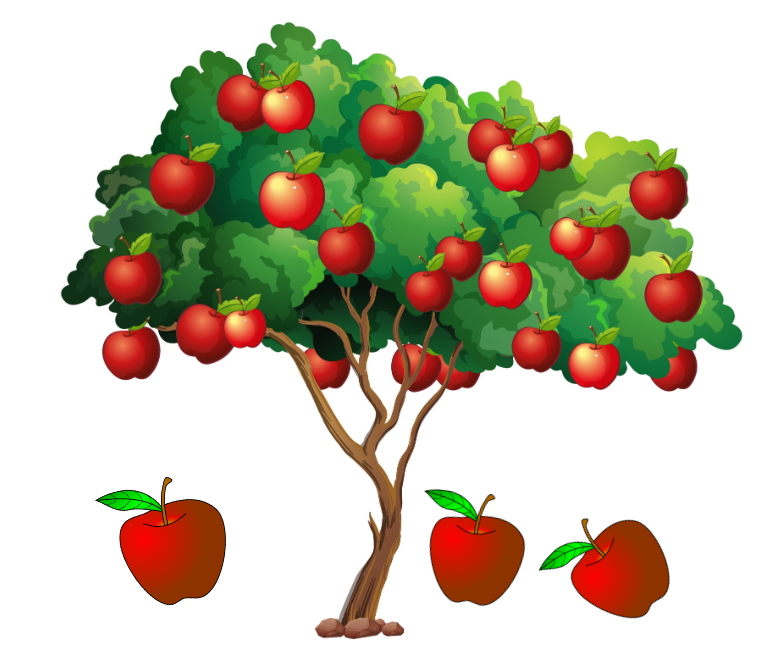}} & \textsc{high}   & \textit{human labeling very useful (e.g. factual correctness)}              \\ \cline{2-3} 
                     & \textsc{medium} & \textit{harder to label (e.g. tone and conciseness)}  \\ \cline{2-3} 
                     & \textsc{low}  & \textit{easy to label with high confidence (e.g. grammatically incorrect)}
\end{tabular}%
}
\caption{Analogy for labeling comparisons: easily labeled comparisons with high quality differential are low hanging fruit and the higher-level fruit requires human annotation}
\label{tab:low_hanging}
\end{table*}
\newpage 
Figure \ref{fig:kendal_rank_graph} presents bar charts illustrating the Kendall Rank correlations between the reward model scores and the SummEval quality measures. Two models are considered for computing these correlations: the checkpoint model represented by the dashed line in Figure \ref{fig:accuracy_graph}, and the `best' model for each specific setting determined based on accuracy. We include the checkpoint results to validate the hypothesis that an earlier model would capture the same features of quality. The corresponding correlation values are detailed in Table \ref{tab:kendal_rank_tab}. Notably, the best values are highlighted in bold, while the worst values are underlined. To provide further context and comparison, we also include \textsc{rouge} scores as additional evaluation metrics.

The high correlation between the reward models and the SummEval scores is a promising indication of their effectiveness. It is worth noting that these reward models outperform all other metrics mentioned in \citet{fabbri2021summeval}, demonstrating their ability to capture quality aspects of summaries beyond superficial features, even when dealing with different domains and varying summary lengths.

We observe that the reward model trained with low agreement comparisons (\textsc{min}) has the lowest correlations for all measures, except for the `consistency' for the checkpoint model.

The reward model in the \textsc{dist} setting demonstrates strong performance, particularly for evaluating coherence and relevance. These dimensions pose notable challenges in measuring summary quality and are characterized by substantial disagreements between experts and non experts, as well as being the poorest measures captured by automatic evaluation, as emphasized in the findings of \citet{fabbri2021summeval}. Therefore, the impressive performance of the \textsc{dist} setting in these categories is a significant accomplishment.

These findings suggest that the reward models learn distinct aspects of summary quality based on the provided comparison data. This raises questions about how we can customise the input data to capture different dimensions of quality effectively.

\subsection{Downstream results}
To assess the impact of our reward models on downstream performance, we incorporate them into the final step of an RLHF framework and evaluate summarization quality using \textsc{rouge} scores. We fine-tune a T5-small model for the summarization task using the quality-filtered TL;DR dataset from \citet{NEURIPS2020_1f89885d}, which consists of 123,169 posts with approximately 5\% held out as a validation set. Figure \ref{fig:overall} presents three graphs displaying the \textsc{rouge-}$1$, \textsc{rouge-}$2$, and \textsc{rouge-}L scores for the downstream models.

Across all three metrics, we observe that integrating the \textsc{dist} reward model yields the highest scores before overfitting occurs. Conversely, the \textsc{min} reward model leads to the poorest model performance when integrated into the downstream process. Interestingly, the \textsc{max} reward model does not demonstrate an increase in \textsc{rouge-}$2$ and \textsc{rouge-}L scores, despite achieving strong quality correlations and accuracy on the comparison test set. 

An interesting finding emerges in terms of downstream reward models: both random selection and distributed agreement models achieve the highest accuracy. This challenges the conventional belief that high agreement should be pursued for reward model paradigms, highlighting the significance of data diversity with respect to choice difficulty in capturing various aspects of quality.

\section{Discussion}


This paper aims to highlight the impact of agreement on the accuracy of reward models and the downstream implications in RLHF. Importantly, the quality differential between choices plays a crucial role in agreement and models trained with examples that elicit different levels of agreement capture distinct aspects of quality. This finding has significant implications for generating synthetic data, as we know that high agreement comparisons occur when there is a high quality differential between the two instances (as demonstrated in Table \ref{tab:comparisons}). Consequently, it becomes possible to generate and automatically label such comparisons, while annotators can focus solely on nuanced and challenging comparisons, resulting in time and cost savings. 

The observation that input data for reward models captures different aspects of quality opens up possibilities for future work in the area of tailored reward modeling. One  avenue for future work is the creation of automatically labeled comparisons to target specific quality characteristics for instance measuring summary factuality. 

The downstream results demonstrate that while high agreement produces good reward models, there is a risk of these models learning superficial characteristics that determine summary quality. Therefore, it is important to consider the diversity of comparison data. Table \ref{tab:low_hanging} contains a visual metaphor to conceptualise this phenomena with respect to summarization. This challenges the conventional belief that maximum agreement always yields optimal results. Our findings indicate that including a range of agreements leads to better downstream performance. However, it is important to note that minimal agreement examples consistently yield the poorest performance in all scenarios. As such, it remains critical to foster good agreement during the annotation process whilst providing room for annotators to express informed disagreements.

\section{Limitations}
One limitation to this work is that we only consider how the reward models perform for the task of summarisation. Extending this work to other natural language generation tasks would require the collection of preference data. To draw more robust conclusions about the impact of agreement for reward models will require more extensive evaluation for different text generation tasks. However, we see this as an initial step in evaluating how agreement impacts multiple facets of the RLHF process. 

In our study, we rely on the dataset collected by \citet{NEURIPS2020_1f89885d} where the coverage of repeat annotations varied. As we were restricted to only considering the samples which has repeat annotations we had a smaller sample of the dataset with which to create our datasets. As such, the distribution of the random dataset is close in nature to the distributed setting. Despite this, there are still pronounced differences when sampling the data for agreement criteria than choosing a random sample. 

The Reddit TL;DR dataset consists of user-submitted posts with minimal moderation and they often contain content that is offensive or reflects harmful social biases. It is important to acknowledge the ethical concerns arising from the use of publicly sourced data without explicit permission from the original parties. While the data we employ is derived from previously released datasets, the content of many posts is highly sensitive and it poses ethical concerns about the use of such data in research. 
\section{Conclusions}
To conclude, this paper highlights the significant impact of training data on the quality aspects represented by reward models and their subsequent downstream performance in RLHF. Contrary to the view that higher agreement is always desirable, our findings demonstrate that training a reward model with a diverse range of agreement levels leads to improved downstream performance. This insight has important implications for data curation and opens up opportunities for synthetic data generation. In future work, we intend to move beyond summarization and extend our work to multiple natural language generation tasks. 
\bibliography{anthology,custom}
\bibliographystyle{acl_natbib}

\end{document}